# Identifying Task Groupings for Multi-Task Learning Using Pointwise V-Usable Information

Yingya Li[a,*], Timothy Miller[a], Steven Bethard[b], Guergana Savova[a]

[a.] Computational Health Informatics Program, Boston Children's Hospital and Harvard Medical School, 401 Park Drive, Boston, MA 02115, USA

[b.] College of Information Science, The University of Arizona, Tucson, Arizona, United States

\* indicates the corresponding author

Corresponding author:
Yingya Li, PhD
Computational Health Informatics Program (CHIP)
Boston Children's Hospital and Harvard Medical School
401 Park Drive, Boston, MA 02115
Email: yingya.li@childrens.harvard.edu

**Funding sources**
The work is funded by grant R01GM114355 from the US National Institutes of Health.

# Abstract

**Objective:** Even in the era of Large Language Models (LLMs) which are claimed to be solutions for many tasks, fine-tuning language models remains a core methodology used in deployment for a variety of reasons – computational efficiency and performance maximization among them. Fine-tuning could be single-task or multi-task joint learning where the tasks support each other thus boosting their performance. The success of multi-task learning can depend heavily on which tasks are grouped together. Naively grouping all tasks or a random set of tasks can result in negative transfer, with the multi-task models performing worse than single-task models. Though many efforts have been made to identify task groupings and to measure the relatedness among different tasks, it remains a challenging research topic to define a metric to identify the best task grouping out of a pool of many potential task combinations. We propose such a metric.

**Methods:** We propose a metric of task relatedness based on task difficulty measured by pointwise $V$-usable information (PVI). PVI is a recently proposed metric to estimate how much usable information a dataset contains given a model. We hypothesize that tasks with not statistically different PVI estimates are similar enough to benefit from the joint learning process. We conduct comprehensive experiments to evaluate the feasibility of this metric for task grouping on 15 NLP datasets in the general, biomedical, and clinical domains. We compare the results of the joint learners against single learners, existing baseline methods, and recent large language models, including Llama and GPT-4.

**Results:** The results show that by grouping tasks with similar PVI estimates, the joint learners yielded competitive results with fewer total parameters, with consistent performance across domains.

**Conclusion:** For domain-specific tasks, finetuned models may remain a preferable option, and the PVI-based method of grouping tasks for multi-task learning could be particularly beneficial. This metric could be wrapped in the overall recipe of fine-tuning language models.

# 1 Introduction

Even in the era of Large Language Models (LLMs) which are claimed to be solutions for many tasks, finetuning language models remains a core methodology used in deployment for a variety of reasons – computational efficiency and performance maximization among them [61]. Finetuning could be single-task or multi-task joint learning (MTL) where the tasks support each other thus boosting their performance.

MTL learns shared representations across tasks and jointly optimizes the losses of all included tasks, which reduces the risk of over-fitting [1]. Compared to single-task learning (STL), MTL has been shown to improve performance and generalization capabilities in many natural language processing (NLP) tasks [2-5]. However, empirical results also suggest that MTL is not always effective and naively grouping tasks brings negative transfer [6-11]. The space of possible task combinations can be massive, and naively searching that space to find the best joint learning models is inefficient [9, 12].

To find the best task combinations, some recent studies have developed new optimization methods that focus on measuring the relatedness among tasks [9, 11-14]. For example, Vu et al. [15] applied task embeddings to predict the transferability of source tasks to a target task. Fifty et al. [9] compared the inter-task affinity by examining how one task's gradient updates on the shared parameters would influence the objective of another task. Song et al. [12] leveraged a meta-learning framework on task combinations. Li et al. [11] applied surrogate models to identify negative transfers among different groupings during MTL and to identify the best combinations for joint learning. However, finding the optimal task grouping usually involves combining many, if not all, tasks for training and optimization, which becomes computationally intensive as the number of tasks increases. For a deeper understanding of the task-relatedness of MTL in neural networks, researchers also provide initial clues to formalize the definition through measurable variables. Specifically, some work suggests that auxiliary tasks with compact and more uniform label distributions are preferable for semantic sequence prediction problems [7]. Others found that gains are more likely to occur for main tasks that plateau quickly with non-plateauing auxiliary tasks [16]. In certain domains like financial NLP tasks, study results show that MTL works well when tasks are related and with diverse skills [13]. Nevertheless, we still lack a shared definition of task-relatedness or a metric to measure the amount of cross-task usable information for a given model under the joint learning context.

This work studies the use of pointwise $\mathcal{V}$-usable information (PVI) [17] to measure the usable information of different datasets and to jointly train tasks with similar information gains given a model. PVI, recently introduced by Ethayarajh et al. [17], estimates the difficulty of data instances for a given model in supervised learning. It builds on the predictive $\mathcal{V}$-information framework [18] which incorporates mutual information and the coefficient of determination to quantify data instance difficulty. The metric applies instance-level predictions to quantify how much information a given model can extract from a dataset. The higher the PVI estimate, the easier it is for the model to represent a given data point. Under this context, we cast PVI as an estimate of task-relatedness to guide MTL. By grouping tasks according to similar PVI distributions, or in other words, tasks of comparable difficulty, we hypothesize that this approach promotes model generalization across the targeted tasks in MTL.

To investigate the feasibility of identifying the best task groupings for MTL using PVI, we conducted experiments with 15 NLP datasets in the general, biomedical and clinical domains. We compared the MTL results with task groupings selected by the PVI estimate distributions against the best-performing fine-tuned single-learner models. The performances were also compared against recent LLMs, including Llama 2 [19], Llama 3 [60], and GPT-4 [20], which have demonstrated their ability as general-purpose NLP task solvers across a wide range of NLP tasks, either with or without downstream data adaptation [21-24]. We also provide a comparison to two baseline task grouping methods: task embedding [15] and surrogate models [11] considered state of the art.

## Statement of Significance

### Problem

In machine learning, naively grouping all tasks or a random set of tasks can result in negative transfer, with the multi-task models performing worse than single-task models.

### What is Already Known

Though many efforts have been made to identify task groupings and to measure the relatedness among different tasks, it remains a challenging research topic to define a metric to identify the best task grouping out of a pool of many potential task combinations.

### What This Paper Adds

We introduce a new method to identify task groupings using PVI estimates for MTL. We provide a thorough empirical analysis across NLP tasks in different domains, demonstrating that our method could effectively find high-performing task groupings that achieve or surpass STL performance. We offer a new perspective on using PVI estimates for task groupings in MTL highlighting its effectiveness for tasks that fall in roughly the same domain.

### Who would benefit from the knowledge in this paper

Researchers working on clinical or general-domain NLP tasks, particularly those who must handle many tasks with limited computing resources, would benefit from the knowledge and method provided in this paper.

## 2 Related Work

### 2.1 MTL with similar tasks

Jointly training similar tasks is the main premise of MTL especially in the era of neural models and transfer learning. The more similar shared tasks are, the more hidden units would be shared in a given model, which would potentially benefit the joint training process through these shared representations [1]. In the context of MTL, measuring task similarity as well as automatically and reliably determining the optimal task grouping from numerous possible configurations remains challenging. Zhang et al. [25] surveyed the efforts in the NLP field for task-relatedness and training methods. Overall, the empirical selection of similar tasks remains the most commonly used method, and in most cases, the problem of deciding which tasks to combine for MTL is often left to human experts [26]. In recent years, a few methods have been developed to automatically select similar tasks [9, 11-15, 27], though a shared definition of task similarity for joint training is still lacking [25]. As a result, negative transfers among tasks in joint training have been observed by researchers [6-7, 9-11, 28]. It would also be computationally costly to search for the best task grouping by iteratively combining all tasks in pairs or n-task groupings. Therefore, a straightforward and efficient method is needed to identify the best task groupings for MTL.

## 2.2 PVI and task difficulty

Understanding the difficulty of a task helps guide the machine learning process, e.g. what architectures and classification methods are feasible [29-31]. The significance of dataset difficulty has been widely discussed in the NLP field [30, 32-34]. As an extension of the predictive V-information framework [18], PVI measures dataset or instance difficulty by the lack of usable information given a model. Algorithm 1 shows how PVI score [17] is calculated (see Section 3 for a detailed description). A high PVI estimate indicates a good representation of the input in the model, and thus the instance is regarded to be easier for the given model. Contrarily, a low PVI estimate indicates the input has less usable information to the model, and the instance is thus considered to be harder. Under the supervised learning context, PVI offers a practical metric that is able to compare the dataset difficulties among different NLP tasks.

PVI has proven effective across downstream tasks, such as quality estimate for universal dependencies [35], informativeness evaluation in the reasoning chain [36], and data augmentation for intent detection [37]. Lu et al. [38] further adapted PVI to in-context learning for LLMs, illustrating its consistency and effectiveness in this new paradigm. In the current study, we propose to explore this metric as a proxy for task similarity to identify the best task groupings for MTL.

# 3 Methods

## 3.1 PVI for task grouping

In this section, we describe our proposed method of utilizing PVI for task groupings in MTL. The method consists of two stages: 1) calculating the PVI estimates for each task; 2) grouping tasks based on PVI score distributions for MTL.

---

**Algorithm 1** The calculation of PVI estimate

---

1. Input: a dataset $D$, a model $G$, a test instance $(x, y) \notin D$
2. $g' \leftarrow$ fine-tune $G$ on $\{(x_i, y_i) | (x_i, y_i) \in D\}$
3. $\emptyset \leftarrow$ empty string
4. $g \leftarrow$ fine-tune $G$ on $\{(\emptyset, y_i) | (x_i, y_i) \in D\}$
5. $PVI(x \rightarrow y) \leftarrow -\log_2 g[\emptyset](y) + \log_2 g'[x](y)$

---

Algorithm 1 shows how the PVI score [17] is calculated where $x_i$ is the input text instance and $y_i$ is its target label. Specifically, it involves fine-tuning a given model $G$ in two different and separate setups indicated by $g'$ and $g$. For $g'$, $G$ is fine-tuned with the input-target pairs $\{(x_i, y_i) | (x_i, y_i) \in D\}$. For $g$, $G$ is fine-tuned only using the targets with empty input text instance, $\{(\emptyset, y_i) | (x_i, y_i) \in D\}$. For each instance in a dataset $D$, given a model $G$, a higher PVI score indicates that the instance $x$ provides more usable information to the model $G$.

Depending on the individual PVI estimates for each instance given a model $G$, our proposed method further groups the datasets by their PVI score distributions. The similarities of distributions over the instances of the datasets are compared using either a paired t-test or a one-way analysis of variance (ANOVA), depending on the number of datasets included with each group.

Datasets with similar PVI score distributions (i.e. not statistically significantly different) are jointly trained in MTL with a hard parameter-sharing strategy, within which the lower layers of a neural network model learn representations across all tasks grouped together, and on top of the shared layers, there are task-specific layers that learn the representations specific to each task

(shown in Figure 1). During model finetuning, we combined datasets from different tasks and applied mini-batch, using a batch size of 32. Each training instance included labels for all tasks, with a pseudo-label of -100 assigned to tasks not relevant to that instance.

Figure 1: An illustration of the architecture for hard parameter sharing of hidden layers in MTL

## 3.2 Experimental setup

### 3.2.1 Datasets and tasks

We conducted experiments on 15 datasets, including 7 general, 2 biomedical, and 6 clinical NLP benchmarks. The details of the datasets, including the data statistics of train, development (dev), and test splits are shown in Tables A.1 and A.2 in the Appendix.

#### 3.2.1.1 General-domain tasks

The general domain tasks include BoolQ [39], CB [40], RTE [41], COPA [42], WiC [43], CoLA [44], SST2 [45]. They are tasks from GLUE [46] and SuperGLUE [41], commonly used as benchmarks for natural language understanding covering question answering, natural language inference, word sense disambiguation, linguistic acceptability, and sentiment analysis.

#### 3.2.1.2 Biomedical tasks

We used the Health Advice [47] and Causal Language [48] datasets representing two fundamental tasks in the biomedical domain: classification and reasoning. Health Advice categorizes biomedical literature sentences into "no advice", "weak advice", and "strong advice" based on recommendation strength. Causal Language identifies correlational and causal statements in biomedical research findings, labeling each as "correlation", "conditional causation", "direct causation" or "no relationship".

#### 3.2.1.3 Clinical tasks

The clinical datasets cover 6 tasks from the THYME corpus (Temporal Histories of Your Medical Events) [49], SHARP Seed (Seed), and SHARP Stratified (Strat). The THYME corpus contains 594 clinical and pathology notes on colon cancer patients, annotated with different syntactic and semantic information. We used the THYME+ version of the corpus [50] and worked with two tasks: negation (THYMENeg) and contextual modality (THYMEMod). Negation indicates if an entity or an event is "negated" or "not negated". Contextual modality flags an entity or an event as "actual" (an event has happened or is scheduled to happen), "hedged/uncertain" (an event is mentioned with some degree of hedging/uncertainty), "hypothetical" (an event is conditional on some other event, usually introduced by "if") or "generic" (an event is mentioned in a general sense, not related to a specific person). The Seed corpus contains notes from patients with pulmonary arterial disease from the Mayo Clinic and patients with breast cancer from Seattle Group Health Cooperative (now part of Keiser Permanante). The Strat corpus contains notes from patients from the Mayo Clinic, with a variety of specialties and note types representing the entire Electronic Medical Records (EMR). We selected the tasks of Negation (SeedNeg, StratNeg) and Uncertainty (SeedUncert, StratUncert) from the two datasets. All of these clinical corpora are de-identified.

### 3.2.2 Models

For the general and biomedical datasets, the base settings for STL and MTL were based on the setting in Liu et al. [51] using RoBERTa-large as the pre-trained model (335M parameters). Though prior work suggests a modest benefit from domain-specific pre-training of the two biomedical tasks [47-48], we used one pre-trained model across the general and biomedical tasks, to allow task combinations across these domains. We provide additional results with bert-base-uncased model (110M parameters) in the Appendix.

Given the unique characteristics of clinical language which is the EMR clinical narrative, we used Bio+Clinical BERT [52] (110M parameters) as the pre-trained model for STL and MTL for the 6 selected clinical tasks. The model was trained on all clinical notes from the MIMIC III dataset [53] atop BioBERT [54] and has been shown to work better for clinical NLP tasks in comparison to other encoder-only BERT-based models [55]. We provide additional results with BioBERT (110M parameters) in the Appendix.

We compared our approach with two baseline task grouping methods: task embedding [15] and surrogate models [11]. The task embedding method effectively identifies transferability across 33 NLP tasks [15] The surrogate model method [11] outperforms other existing task grouping strategies in both NLP and computer vision tasks, such as HOA [56] and TAG [9]. We also compared the proposed method with two additional baselines: (1) joint training of all tasks together, and (2) joint training using randomly selected pairs of tasks.

To further evaluate the MTL performance using PVI estimates, we compared the results with recent LLMs including LlaMA2-7B-chat, LlaMA2-70B-chat [19], Llama-3.2-3B-Instruct [60], and GPT-4 [20], using few-shot prompting with the number of shots equal to the number of classes in each task. This comparison aims to determine if LLMs' performance could match or exceed our best MTL models, as LLMs have been claimed to handle various NLP tasks with impressive performance even without downstream adaptation and thus could be considered a type of multi-task learners. Following prior studies in prompting design [38], we designed prompts (see Tables A.16 and A.17 in Appendix for examples). We limited the evaluation of GPT-4 to experiments on the 7 general and 2 biomedical datasets, due to the HIPAA (Health Insurance Portability and Accountability Act) regulation of the clinical datasets[1].

### 3.2.3 Setting and evaluation

To obtain the PVI estimates for all datasets, we fine-tuned the base models for 10 epochs with a batch size of 32, and a learning rate of $2e$-5 through HuggingFace's Transformer API [57]. Unless specified, all other settings in our experiments are the default ones.

For STL and MTL with different task groupings, we fine-tuned the models with different seed values, epoch numbers, and learning rates. More details of the experiments are in the first paragraph of the Appendix. In consideration of the skewed label distribution for some of the tasks, we adopted both accuracy and macro F1-score as our evaluation metrics. All the datasets were separated into train, dev and test sets. Results for all tasks, using different models and settings are reported on the test set except where the test split labels were not publicly available[2].

## 4 Results

We conducted paired t-tests to compare the PVI estimates of pairs of tasks (group size = 2) using the same model. This analysis yielded 36 task groupings for the general and biomedical domains and 15 groupings for the clinical domain. Table 1A shows the results of task groupings where there is no statistically significant difference in PVI estimates between the grouped tasks ($p$-value > 0.01) and therefore the two tasks have a similar level of difficulty for the given model.

---

[1] We do not have access to the HIPAA-compliant GPT-4 and are not allowed to pass the data to the public GPT-4 API.
[2] GLUE and SuperGLUE tasks have no ground truth labels in the original test set. We divided the original training set into a train and a dev split. Like prior studies [58-59], the original dev set serves as the held-out, eyes-off dataset, which was used only to report the results in this paper.

For task groupings with more than 2 tasks (group size > 2), we used one-way ANOVA to examine the difference in PVI estimates across different tasks. Table 1B shows the results of task groupings with similar PVI estimates distribution ($p$-value > 0.01). We also observed that in group sizes larger than 3, no tasks have similar difficulty levels, as measured by the one-way ANOVA of PVI estimates. This means that among all tasks we experimented with, no more than three tasks are equally challenging for the pre-trained model, i.e. no more than three tasks would fit our criteria for task grouping.

| A: | | |
| --- | --- | --- |
| **Task Group** | *T*-statistic | *P*-value |
| Health Advice-Causal Language | 1.398 | 0.162 |
| Causal-CB | 2.411 | 0.016 |
| BoolQ-RTE | 1.775 | 0.076 |
| CB-COPA | 1.328 | 0.185 |
| CB-CoLA | 1.237 | 0.216 |
| CB-SST2 | -1.660 | 0.097 |
| WiC-COPA | -2.358 | 0.019 |
| COPA-CoLA | -0.072 | 0.943 |
| THYMENeg-SeedNeg | 1.680 | 0.093 |
| THYMEMod-SeedNeg | -1.740 | 0.082 |
| THYMEMod-StratNeg | 0.585 | 0.559 |
| SeedNeg-StratNeg | 2.546 | 0.011 |
| SeedUncert-StratUncert | 0.380 | 0.704 |
| **B:** | | |
| **Task Grouping** | *F*-statistic | *P*-value |
| CB-CoLA-COPA | 0.813 | 0.444 |
| THYMEMod-SeedUncert-StratNeg | 4.330 | 0.013 |
| THYMEMod-SeedNeg-StratNeg | 1.801 | 0.163 |

Table 1: Task groupings with PVI estimates that are not significantly different based on statistical measurements.

## 4.1 MTL results of grouped tasks with similar PVI estimates

To test the feasibility of utilizing task difficulty for task grouping, we jointly trained on the datasets with similar PVI estimates. Tables 2 and 3 show the accuracy and macro F1 results of the single-task learners and joint learning models when combining tasks with similar estimates.[3] For tasks that have the PVI estimate similar to more than one task, such as CB, as shown in Table 1A, we selected the grouping with larger *T*-statistic values for the joint training.

| Task Group | STL (acc.) | STL (f1) | MTL (acc.) | MTL (f1) |
|---|---|---|---|---|
| BoolQ | 0.806 | 0.751 | 0.854 | 0.843 |
| | ±0.128 | ±0.255 | ±0.003 | ±0.004 |
| RTE | 0.829 | 0.828 | 0.842 | 0.841 |
| | ±0.011 | ±0.010 | ±0.015 | ±0.014 |
| CB | 0.811 | 0.650 | 0.925 | 0.907 |
| | ±0.073 | ±0.192 | ±0.039 | ±0.056 |
| Causal Language | 0.862 | 0.825 | 0.862 | 0.825 |
| | ±0.008 | ±0.010 | ±0.008 | ±0.008 |
| COPA | 0.580 | 0.552 | 0.657 | 0.654 |
| | ±0.076 | ±0.124 | ±0.049 | ±0.049 |
| WiC | 0.695 | 0.689 | 0.686 | 0.679 |
| | ±0.012 | ±0.015 | ±0.017 | ±0.021 |
| CoLA | 0.844 | 0.800 | 0.846 | 0.802 |
| | ±0.020 | ±0.029 | ±0.020 | ±0.027 |
| COPA | 0.580 | 0.552 | 0.678 | 0.641 |
| | ±0.076 | ±0.124 | ±0.139 | ±0.224 |
| SST2 | 0.961 | 0.961 | 0.961 | 0.961 |
| | ±0.002 | ±0.002 | ±0.002 | ±0.002 |
| CB | 0.811 | 0.650 | 0.932 | 0.903 |
| | ±0.073 | ±0.192 | ±0.063 | ±0.100 |
| Health Advice | 0.939 | 0.934 | 0.937 | 0.930 |
| | ±0.005 | ±0.006 | ±0.005 | ±0.005 |
| Causal Language | 0.862 | 0.825 | 0.863 | 0.827 |
| | ±0.008 | ±0.010 | ±0.018 | ±0.020 |
| CB | 0.811 | 0.650 | 0.904 | 0.886 |
| | ±0.073 | ±0.192 | ±0.040 | ±0.054 |
| CoLA | 0.844 | 0.800 | 0.812 | 0.719 |
| | ±0.020 | ±0.029 | ±0.085 | ±0.217 |
| COPA | 0.580 | 0.552 | 0.836 | 0.790 |
| | ±0.076 | ±0.124 | ±0.012 | ±0.018 |

Table 2: A comparison of performance between STL and MTL by task groupings with similar PVI estimates for the general and biomedical tasks on the test dataset. Both STL and MTL were fine-tuned using roberta-large.

---

[3] In MTL, datasets are jointly trained in pairs (group size = 2) or sets of three (group size = 3). For example, BoolQ and RTE in the first row were jointly trained. MTL scores (accuracy and macro-F1) indicate performance for each dataset in a group. Average results were reported with a 95% confidence interval from 5 random seed samples (t = 2.776).

| Task Group | STL (acc.) | STL (f1) | MTL (acc.) | MTL (f1) |
|---|---|---|---|---|
| THYMEMod | 0.957 | 0.736 | 0.958 | 0.743 |
| | ±0.001 | ±0.009 | ±0.001 | ±0.007 |
| SeedNeg | 0.983 | 0.944 | 0.986 | 0.955 |
| | ±0.002 | ±0.008 | ±0.002 | ±0.005 |
| THYMENeg | 0.983 | 0.954 | 0.984 | 0.956 |
| | ±0.001 | ±0.003 | ±0.001 | ±0.002 |
| SeedNeg | 0.983 | 0.944 | 0.986 | 0.955 |
| | ±0.002 | ±0.008 | ±0.004 | ±0.012 |
| SeedUncert | 0.973 | 0.829 | 0.977 | 0.859 |
| | ±0.004 | ±0.031 | ±0.006 | ±0.034 |
| StratUncert | 0.982 | 0.675 | 0.987 | 0.845 |
| | ±0.002 | ±0.071 | ±0.002 | ±0.027 |
| StratNeg | 0.987 | 0.925 | 0.989 | 0.941 |
| | ±0.002 | ±0.012 | ±0.001 | ±0.004 |
| SeedNeg | 0.983 | 0.944 | 0.984 | 0.949 |
| | ±0.002 | ±0.008 | ±0.002 | ±0.005 |
| THYMEMod | 0.957 | 0.736 | 0.958 | 0.736 |
| | ±0.001 | ±0.009 | ±0.001 | ±0.012 |
| SeedUncert | 0.973 | 0.829 | 0.979 | 0.868 |
| | ±0.004 | ±0.031 | ±0.005 | ±0.029 |
| StratNeg | 0.987 | 0.925 | 0.985 | 0.915 |
| | ±0.006 | ±0.009 | ±0.003 | ±0.015 |
| THYMEMod | 0.957 | 0.736 | 0.958 | 0.741 |
| | ±0.001 | ±0.009 | ±0.002 | ±0.012 |
| SeedNeg | 0.983 | 0.944 | 0.983 | 0.944 |
| | ±0.002 | ±0.008 | ±0.002 | ±0.008 |
| StratNeg | 0.987 | 0.925 | 0.987 | 0.929 |
| | ±0.002 | ±0.012 | ±0.001 | ±0.005 |

Table 3: A comparison of performance between STL and MTL by task groupings with similar PVI estimates for the clinical tasks on the test dataset. Both STL and MTL were fine-tuned using Bio+Clinical BERT.

For all general and biomedical tasks the joint learning (MTL) results when grouping two tasks with statistically similar PVI estimates are either similar or better than the STL (margin of error factored in). Based on the accuracy and F1 scores, for the 9 general and biomedical tasks, MTL for groupings (group size = 2) of statistically similar PVI estimates produced results that were either comparable to or better than those of the single learners for all datasets. For WiC, Health Advice, and CoLA when paired with COPA, Causal Language, CB and COPA the MTL results are within the margin of error when compared to the STL results. The CB task has the largest F1-score improvement (**Δ** = 0.257). Similar patterns are also observed with grouping sizes larger than 2 when 3 tasks with similar PVI estimates are combined for joint training.

Results of the clinical tasks are in Table 3. The observed consistent improvement of PVI-based MTL grouping as compared to single learners may be due to the more constrained and formulaic nature of the language in this field, which better supports task semantics than the broader variability seen in the general domain. The same trends are exhibited when we experimented with other language models with different architectures (BERT-base-uncased, BioBERT) shown in Tables A.7 (for the general and biomedical tasks) and A.8 (for the clinical tasks) in Appendix. To further examine the methods, we conducted an additional small-scale experiment (due to computational power resource constraints) in which we fine-tuned the Llama-3.2-1B model with LoRA on six general-domain tasks (BoolQ, RTE, CB, CoLA, WiC and COPA) to compare the performance.. The groupings were based on the PVI scores (as listed in Table A.9). Compared to BERT-based models, the PVI-based groupings are the same for BoolQ-RTE and WiC-COPA but different for CB-CoLA as the additional small-scale experiment included a much smaller task set and a different model. The results (as shown in Table A.10) show similar patterns to those observed with the BERT-based models, with most tasks showing improved average accuracy and F1 scores under joint learning (MTL). Although RTE and CoLA exhibit higher mean performance in the single-task setting, the substantial overlap in their 95% confidence intervals suggests these differences likely reflect random variation rather than systematic performance drop when training tasks jointly.

Compared with the best MTL learning results of the two baseline approaches, shown in Tables A.3-A.6 in Appendix, our method of utilizing similar PVI estimates also shows either the same (e.g., Health advice and Causal Language) or a large improvement among the tasks in different domains. Results for THYMENeg combined with THYMEMod are similar within the margin of error. Our method introduces performance improvements when compared to the recent SOTA MTL grouping method – the surrogate model method [11] – which is reported to outperform other existing task grouping strategies on both NLP and computer vision tasks, such as HOA [56] and TAG [9]. Two additional baseline approaches (shown in Tables A.11–A.14 in the Appendix) perform worse than our proposed task grouping method. In particular, grouping all tasks together leads to significant performance drops across most tasks – especially in the general and biomedical domains – compared to both single-task fine-tuning and our method.

## 4.2 MTL results of grouped tasks with different PVI estimates

To further examine how task difficulty may affect the performance of joint learning when grouping tasks by their PVI estimates, we combined tasks with the most statistically different PVI estimates ($p$-value < 0.01) for joint learning. Tables 4 and 5 show the results on the general, biomedical, and clinical tasks individually. Compared to STL and MTL with task grouping of similar PVI estimates, the performance is much lower. Even though the CB dataset showed a better performance when combined with the Health Advice datasets, it came at a cost for the other task - the performance of Health Advice was lower compared to its single learner result. On the other hand, the results of the Health Advice dataset are similar to those from MTL with similar PVI grouping however it comes at a cost for the other task (as shown in Table 4).

| Task Group | STL (acc.) | STL (f1) | MTL (acc.) | MTL (f1) |
|---|---|---|---|---|
| BoolQ | 0.806 ±0.128 | 0.751 ±0.255 | 0.787 ±0.115 | 0.729 ±0.240 |
| Health Advice | 0.939 ±0.005 | 0.934 ±0.006 | 0.936 ±0.004 | 0.929 ±0.005 |
| CB | 0.811 ±0.073 | 0.650 ±0.192 | 0.846 ±0.046 | 0.664 ±0.110 |
| Health Advice | 0.939 ±0.005 | 0.934 ±0.006 | 0.936 ±0.006 | 0.926 ±0.008 |
| RTE | 0.829 ±0.011 | 0.828 ±0.010 | 0.734 ±0.134 | 0.714 ±0.182 |
| Health Advice | 0.939 ±0.005 | 0.934 ±0.006 | 0.938 ±0.010 | 0.931 ±0.013 |
| COPA | 0.580 ±0.076 | 0.552 ±0.124 | 0.514 ±0.046 | 0.453 ±0.105 |
| Health Advice | 0.939 ±0.005 | 0.934 ±0.006 | 0.938 ±0.005 | 0.933 ±0.007 |
| WiC | 0.695 ±0.012 | 0.689 ±0.015 | 0.674 ±0.026 | 0.667 ±0.029 |
| Health Advice | 0.939 ±0.005 | 0.934 ±0.006 | 0.939 ±0.003 | 0.933 ±0.003 |
| CoLA | 0.844 ±0.020 | 0.800 ±0.029 | 0.844 ±0.006 | 0.798 ±0.008 |
| Health Advice | 0.939 ±0.005 | 0.934 ±0.006 | 0.939 ±0.003 | 0.932 ±0.006 |
| SST2 | 0.961 ±0.002 | 0.961 ±0.002 | 0.953 ±0.007 | 0.953 ±0.007 |
| Health Advice | 0.939 ±0.005 | 0.934 ±0.006 | 0.929 ±0.029 | 0.917 ±0.045 |
| Causal Language | 0.862 ±0.008 | 0.825 ±0.010 | 0.863 ±0.012 | 0.825 ±0.015 |
| BoolQ | 0.806 ±0.128 | 0.751 ±0.255 | 0.826 ±0.005 | 0.812 ±0.006 |

Table 4: Performance of MTL by task groupings with different PVI estimates for the general and biomedical datasets. Both STL and MTL were fine-tuned using roberta-large.

| Task Group | STL (acc.) | STL (f1) | MTL (acc.) | MTL (f1) |
|---|---|---|---|---|
| THYMEMod | 0.957 ±0.001 | 0.736 ±0.009 | 0.956 ±0.002 | 0.730 ±0.008 |
| StratUncert | 0.982 ±0.002 | 0.675 ±0.071 | 0.984 ±0.003 | 0.813 ±0.030 |
| THYMENeg | 0.983 ±0.001 | 0.954 ±0.003 | 0.984 ±0.001 | 0.957 ±0.003 |
| THYMEMod | 0.957 ±0.001 | 0.736 ±0.009 | 0.958 ±0.001 | 0.740 ±0.008 |
| SeedNeg | 0.938 ±0.002 | 0.944 ±0.008 | 0.979 ±0.003 | 0.936 ±0.008 |
| StratUncert | 0.982 ±0.002 | 0.675 ±0.071 | 0.982 ±0.004 | 0.671 ±0.108 |
| SeedUncert | 0.973 ±0.004 | 0.829 ±0.031 | 0.972 ±0.008 | 0.795 ±0.068 |
| StratNeg | 0.987 ±0.002 | 0.925 ±0.012 | 0.985 ±0.002 | 0.919 ±0.013 |
| StratNeg | 0.987 ±0.002 | 0.925 ±0.012 | 0.986 ±0.001 | 0.919 ±0.007 |
| THYMENeg | 0.983 ±0.001 | 0.954 ±0.003 | 0.983 ±0.001 | 0.953 ±0.002 |
| StratUncert | 0.982 ±0.002 | 0.675 ±0.071 | 0.981 ±0.001 | 0.636 ±0.033 |
| THYMENeg | 0.983 ±0.001 | 0.954 ±0.003 | 0.983 ±0.001 | 0.953 ±0.002 |

Table 5: Performance of MTL by task groupings with different PVI estimates for the clinical datasets. Both STL and MTL were fine-tuned using Bio+Clinical BERT.

We observed a similar pattern for the tasks in the clinical domain, except for the slight improvement over STL when THYMEMod is jointly trained with THYMENeg, and a better performance of StratUncert when combined with THYMEMod (at cost for THYMEMod). This result concurs with previous findings that naively grouping tasks for joint training brings negative transfer among the tasks, thus leading to the incorrect conclusion that MTL does not work.

### 4.3 Performance of LLMs on the tasks

LLMs are intended to provide solutions for many tasks, thus are considered multi-task learners. Therefore, we compare our approach to the performance of LLMs as a baseline. Table 6 shows the performance of Llama2-7B, Llama2-70B, Llama3.2-3B, and GPT-4 on the general and biomedical tasks using few-shot learning. Overall, our best STL and MTL models performed better than the Llama models for all tasks. On the other hand, GPT-4 shows mixed results in terms of performance compared to the best STL and MTL. It outperformed the best MTL results on RTE, CB, COPA, CoLA and SST2 but underperformed on BoolQ and WiC. It is unclear to what extent these tasks are included in the training corpus of GPT-4. As for the two biomedical tasks, our best MTL model outperformed GPT-4 by a wide margin (0.234-0.331 F1 points), with much higher performance with MTL with task groupings with a similar PVI-based difficulty given the roberta-large model.

Similar to the biomedical datasets, the STL and MTL models of the clinical tasks outperformed the Llama models, which indicates that for domain-specific NLP tasks, fine-tuning tasks-specific transformer-based models still works better than prompting the LLMs. Moreover, EMR text is highly unlikely to have been included in the training data of any model given the HIPAA provisions, thus it is ideal for independent evaluations of LLMs. Note, that we do not have access to HIPAA-compliant GPT models, thus do not provide results with these models on the clinical tasks. We believe that the trends exhibited in the general and biomedical tasks regarding comparison of GPT-4 and Llama models are likely to hold for the clinical tasks as well.

The comparison of Llama2-7B v. Llama2-70B (Table 6) and roberta-large vs. bert-base-uncased (Appendix Table A.3 and A.7) demonstrates that the size of the model matters – models with more parameters yield better results.

| Dataset | Model | Acc. | F1 |
|---|---|---|---|
| BoolQ | Llama2-7B (few-shot) | 0.507 | 0.309 |
| | Llama2-70B (few-shot) | 0.625 | 0.378 |
| | Llama3-3B (few-shot) | 0.478 | 0.477 |
| | GPT-4 (few-shot) | 0.697 | 0.591 |
| | Ours (fine-tuned) | 0.854 ±0.003 | 0.843 ±0.004 |
| CB | Llama2-7B (few-shot) | 0.250 | 0.258 |
| | Llama2-70B (few-shot) | 0.446 | 0.323 |
| | Llama3-3B (few-shot) | 0.464 | 0.415 |
| | GPT-4 (few-shot) | 0.964 | 0.932 |
| | Ours (fine-tuned) | 0.925 ±0.039 | 0.907 ±0.056 |
| RTE | Llama2-7B (few-shot) | 0.498 | 0.332 |
| | Llama2-70B (few-shot) | 0.783 | 0.522 |
| | Llama3-3B (few-shot) | 0.574 | 0.356 |
| | GPT-4 (few-shot) | 0.895 | 0.894 |
| | Ours (fine-tuned) | 0.842 ±0.015 | 0.841 ±0.014 |
| COPA | Llama2-7B (few-shot) | 0.450 | 0.227 |
| | Llama2-70B (few-shot) | 0.495 | 0.331 |
| | Llama3-3B (few-shot) | 0.540 | 0.533 |
| | GPT-4 (few-shot) | 0.940 | 0.940 |
| | Ours (fine-tuned) | 0.836 ±0.012 | 0.790 ±0.018 |
| WiC | Llama2-7B (few-shot) | 0.517 | 0.337 |
| | Llama2-70B (few-shot) | 0.586 | 0.576 |
| | Llama3-3B (few-shot) | 0.500 | 0.250 |
| | GPT-4 (few-shot) | 0.502 | 0.358 |
| | Ours (fine-tuned) | 0.686 ±0.017 | 0.679 ±0.021 |
| CoLA | Llama2-7B (few-shot) | 0.249 | 0.178 |
| | Llama2-70B (few-shot) | 0.740 | 0.646 |
| | Llama3-3B (few-shot) | 0.438 | 0.427 |
| | GPT-4 (few-shot) | 0.856 | 0.829 |
| | Ours (fine-tuned) | 0.846 ±0.020 | 0.802 ±0.027 |
| SST2 | Llama2-7B (few-shot) | 0.232 | 0.179 |
| | Llama2-70B (few-shot) | 0.930 | 0.930 |
| | Llama3-3B (few-shot) | 0.536 | 0.420 |
| | GPT-4 (few-shot) | 0.964 | 0.964 |
| | Ours (fine-tuned) | 0.961 ±0.002 | 0.961 ±0.002 |
| Health Advice | Llama2-7B (few-shot) | 0.316 | 0.213 |
| | Llama2-70B (few-shot) | 0.205 | 0.185 |
| | Llama3-3B (few-shot) | 0.317 | 0.317 |
| | GPT-4 (few-shot) | 0.719 | 0.700 |
| | Ours (fine-tuned) | 0.937 ±0.005 | 0.930 ±0.005 |
| Causal Language | Llama2-7B (few-shot) | 0.165 | 0.113 |
| | Llama2-70B (few-shot) | 0.282 | 0.255 |
| | Llama3-3B (few-shot) | 0.227 | 0.212 |
| | GPT-4 (few-shot) | 0.545 | 0.494 |
| | Ours (fine-tuned) | 0.863 ±0.018 | 0.827 ±0.020 |
| THYMENeg | Llama2-7B (few-shot) | 0.309 | 0.196 |
| | Llama2-70B (few-shot) | 0.379 | 0.342 |
| | Llama3-3B (few-shot) | 0.898 | 0.473 |
| | Ours (fine-tuned) | 0.984 ±0.001 | 0.956 ±0.002 |
| THYMEMod | Llama2-7B (few-shot) | 0.673 | 0.184 |
| | Llama2-70B (few-shot) | 0.177 | 0.073 |
| | Llama3-3B (few-shot) | 0.471 | 0.119 |
| | Ours (fine-tuned) | 0.958 ±0.001 | 0.743 ±0.007 |
| SeedNeg | Llama2-7B (few-shot) | 0.462 | 0.246 |
| | Llama2-70B (few-shot) | 0.618 | 0.428 |
| | Llama3-3B (few-shot) | 0.355 | 0.316 |
| | Ours (fine-tuned) | 0.986 ±0.002 | 0.955 ±0.005 |
| SeedUncert | Llama2-7B (few-shot) | 0.738 | 0.299 |
| | Llama2-70B (few-shot) | 0.204 | 0.126 |
| | Llama3-3B (few-shot) | 0.864 | 0.503 |
| | Ours (fine-tuned) | 0.979 ±0.005 | 0.868 ±0.029 |
| StratNeg | Llama2-7B (few-shot) | 0.226 | 0.139 |
| | Llama2-70B (few-shot) | 0.487 | 0.369 |
| | Llama3-3B (few-shot) | 0.657 | 0.426 |
| | Ours (fine-tuned) | 0.989 ±0.001 | 0.941 ±0.004 |
| StratUncert | Llama2-7B (few-shot) | 0.397 | 0.204 |
| | Llama2-70B (few-shot) | 0.690 | 0.420 |
| | Llama3-3B (few-shot) | 0.877 | 0.476 |
| | Ours (fine-tuned) | 0.987 ±0.002 | 0.845 ±0.027 |

Table 6: Performance of Llama2-7B, Llama2-70B, Llama3.2-3B, and GPT-4 on the general, biomedical, and clinical tasks.

## 5 Conclusion

In this study, we propose a new method of identifying best-performing task groupings for MTL based on PVI estimates. We conducted experiments on 15 NLP datasets in the general, biomedical, and clinical domains. We showed that when grouping tasks with similar (i.e. not significantly different) PVI estimates, MTL yielded competitive or better results on the majority of the general domain tasks and all biomedical and clinical tasks included in our study. The performance was much better compared to the task grouping with PVI estimates of datasets that are significantly different, suggesting the importance of considering task difficulty as task relatedness and the feasibility of utilizing PVI as a metric for selecting task combinations.

Though MTL may lead to only minor improvements for some tasks as compared to STL, it is more efficient than STL, given the reduced need for training and deployment of multiple models within the computational environment of the average institution, e.g. academic medical center or hospital. For instance, applying two separate roberta-large models for the two biomedical tasks in our experiment would require tuning ~710 million parameters (355 million for each model) as listed in Table A.15 in the Appendix. In contrast, MTL with hard parameter sharing involves tuning the shared 355 million parameters once, plus a small percentage of task-specific parameters, with the total number of parameters significantly less than 710 million. While prompting LLMs yielded varying results on the general domain datasets, both STL and MTL models consistently outperformed Llama and GPT-4 models in the biomedical domain. This indicates that for domain-specific tasks, fine-tuned models may remain a preferable option. Due to the computational capacity needed for fine-tuning LLMs and the data privacy regulation when using the LLMs, the PVI-based method of grouping tasks for MTL could be particularly beneficial for domain-specific tasks.

For the PVI grouping results, we observed that in the general domain, tasks such as CoLA and WiC showed smaller gains or near-equivalent performance under multitask learning (as shown in Table 2), possibly due to their unique linguistic characteristics and label distributions, which are less aligned with those of other tasks in their group. In contrast, in the biomedical and clinical tasks, such as SeedUncert and StratUncert, multitask learning consistently improved F1 scores by substantial margins compared to single-task learning (as shown in Table 3), suggesting benefits when tasks share similar label schemes and domain-specific terminology. These examples highlight that while PVI provides a valuable signal for grouping, its effectiveness can be further enhanced by incorporating complementary measures of task similarity, such as vocabulary overlap or label distribution divergence.

This study focuses on task grouping for MTL using PVI. This metric could also be applied for data instance selection to optimize training sets for joint learning. Moreover, investigating parameter sharing and monitoring the variation in weights for both single learners and joint learning models, by ranking parameters based on their change after grouping tasks, represents another avenue for improved MTL's effectiveness and efficiency. We will explore these directions in future work. We believe our proposed approach could be incorporated into frameworks for stacking LMs to algorithmically optimize LM prompting, fine-tuning, augmentation, and reasoning.

As noted, one limitation of the current study is that we were not able to test GPT-4 performance on the clinical tasks because we do not have access to a HIPAA-compliant GPT-4 model and are not allowed to transmit data to the GPT4 public API. We chose to experiment with two of the most competitive LLMs — the Llama family of models which are locally downloadable and GPT-4. Experiments with other LLMs can be pursued, however we believe the LLMs results we report in this study are likely representative of the general trends.

The results of the single learners are with smaller models (less than 7B parameters) as fine-tuning and PVI estimates are computationally feasible given our computational resources. We encourage those with more abundant computational resources to experiment with our methodology using LLMs as the base models which we believe will further improve our reported results. We believe that including our method for multi-task grouping in the recipe for fine-tuning multi-task language models is a feasible path.

The clinical datasets we worked with represent a limited dataset where all confidential data are removed except for dates. We did not transmit any part of these clinical datasets to any public API and processed the dataset locally on a HIPAA-compliant server. We have an approved IRB for the study described in this paper.

**Data availability statement**

The general and biomedical datasets are publicly available via the published work cited in the references.

**Author contributions**

YL: conceptualization, data curation, formal analysis, investigation, methodology, visualization, writing - original draft, writing - review & editing

TM: conceptualization, data curation, formal analysis, investigation, methodology, writing - review & editing

SB: conceptualization, data curation, formal analysis, investigation, methodology, writing - original draft, writing - review & editing

GS: conceptualization, data curation, formal analysis, investigation, methodology, supervision, visualization, funding acquisition, resources, writing - original draft, writing - review & editing

**Competing interests**

Authors YL, TM, SB, and GS declare no competing interests.

**Acknowledgment**

The study was funded by R01GM114355 (Authors YL, GS, SB, TM). The funder played no role in the study design, data collection, analysis and interpretation of data, or the writing of this manuscript.

# Appendix

We used an NVIDIA A40 GPU cluster of 4 nodes for the single- and multi-task training experiments. We experimented with training epochs of {5, 6, 7, 8, 9, 10}, random seeds of {42, 52, 62, 72, 82}, a max sequence length of {126, 512}, a learning rate of {1e-5, 2e-5}, with a fixed batch size of 32 and gradient accumulation steps of 2. All experiments with LLMs were run using an NVIDIA A100 or through OpenAI API. Table A.1 provides details of the datasets we used. Table A.2 shows the label distribution of each dataset.

| Dataset | Domain | Train \| Dev \| Test Sets Size (# instances) | Type of Task | # Classes |
|---|---|---|---|---|
| BoolQ | General | 7541 \| 1886 \| 3270 | Question and Answering | 2 |
| CB | General | 200 \| 50 \| 57 | Natural Language Inference | 3 |
| RTE | General | 1992 \| 498 \| 277 | Text Entailment | 2 |
| COPA | General | 320 \| 80 \| 100 | Question and Answering | 2 |
| WiC | General | 4342 \| 1086 \| 638 | Word Sentence Disambiguation | 2 |
| CoLA | General | 8640 \| 1711 \| 1043 | Acceptability | 2 |
| SST2 | General | 53879 \| 13470 \| 872 | Sentiment Analysis | 2 |
| Health Advice | Biomedical | 3828 \| 957 \| 1197 | Suggestion Mining | 3 |
| Causal Language | Biomedical | 1958 \| 490 \| 614 | Causal Relation | 4 |
| THYMENeg | Clinical | 28637 \| 7160 \| 19240 | Negation Detection | 2 |
| THYMEMod | Clinical | 28637 \| 7160 \| 19240 | Contextual Modality Detection | 4 |
| SeedNeg | Clinical | 4666 \| 1167 \| 676 | Negation Detection | 2 |
| SeedUncert | Clinical | 4666 \| 1167 \| 676 | Uncertainty Detection | 2 |
| StratNeg | Clinical | 3568 \| 802 \| 1737 | Negation Detection | 2 |
| StratUncert | Clinical | 3568 \| 802 \| 1737 | Uncertainty Detection | 2 |

Table A.1: Overview of datasets used in this study. The datasets cover a variety of tasks and domains, providing a comprehensive base for our analyses.

| Task | Label | Train | Dev | Test |
|---|---|---|---|---|
| BoolQ | True | 4699 | 1175 | 2033 |
|  | False | 2842 | 711 | 1237 |
| CB | Contradiction | 95 | 24 | 28 |
|  | Entailment | 92 | 23 | 23 |
|  | Neutral | 13 | 3 | 5 |
| RTE | Entailment | 999 | 250 | 146 |
|  | Not Entailment | 993 | 248 | 131 |
| COPA | 0 | 160 | 40 | 50 |
|  | 1 | 160 | 40 | 50 |
| WiC | True | 2171 | 543 | 319 |
|  | False | 2171 | 543 | 319 |
| CoLA | 0 | 2022 | 506 | 322 |
|  | 1 | 4818 | 1205 | 721 |
| SST2 | 0 | 23824 | 5956 | 428 |
|  | 1 | 30055 | 7514 | 444 |
| Health Advice | No Advice | 2296 | 574 | 705 |
|  | Weak Advice | 943 | 236 | 303 |
|  | Strong Advice | 589 | 147 | 189 |
| Causal Language | No Relation | 860 | 215 | 281 |
|  | Correlation | 639 | 160 | 199 |
|  | Conditional Causation | 320 | 80 | 94 |
|  | Direct Causation | 139 | 35 | 39 |
| THYMENeg | 0 | 3079 | 770 | 1961 |
|  | 1 | 25558 | 6390 | 17279 |
| THYMEMod | Actual | 26358 | 6590 | 17750 |
|  | Hypothetical | 1236 | 309 | 781 |
|  | Hedged | 656 | 164 | 417 |
|  | Generic | 387 | 97 | 292 |
| SeedNeg | 0 | 3960 | 991 | 619 |
|  | 1 | 704 | 176 | 57 |
| SeedUncert | 0 | 4466 | 1117 | 647 |
|  | 1 | 198 | 50 | 29 |
| StratNeg | 0 | 3033 | 759 | 1658 |
|  | 1 | 173 | 43 | 80 |
| StratUncert | 0 | 3118 | 780 | 1700 |
|  | 1 | 88 | 22 | 38 |

Table A.2: Label distribution of each datasets.

| Task Group | Ours (Acc.) | Ours (F1) | TaskEmb (Acc.) | TaskEmb (F1) |
|---|---|---|---|---|
| BoolQ | 0.854 | 0.843 | 0.830 | 0.817 |
|  | ±0.003 | ±0.004 | ±0.005 | ±0.005 |
| CB | 0.925 | 0.907 | 0.925 | 0.899 |
|  | ±0.039 | ±0.056 | ±0.053 | ±0.090 |
| RTE | 0.842 | 0.841 | 0.588 | 0.520 |
|  | ±0.015 | ±0.014 | ±0.147 | ±0.237 |
| SST2 | 0.961 | 0.961 | 0.775 | 0.706 |
|  | ±0.002 | ±0.002 | ±0.302 | ±0.419 |
| COPA | 0.836 | 0.790 | 0.505 | 0.365 |
|  | ±0.012 | ±0.018 | ±0.014 | ±0.090 |
| WiC | 0.686 | 0.679 | 0.540 | 0.433 |
|  | ±0.017 | ±0.021 | ±0.076 | ±0.175 |
| CoLA | 0.846 | 0.802 | 0.802 | 0.700 |
|  | ±0.020 | ±0.027 | ±0.083 | ±0.209 |
| SST2 | 0.961 | 0.961 | 0.862 | 0.827 |
|  | ±0.002 | ±0.002 | ±0.245 | ±0.341 |
| Health Advice | 0.937 | 0.930 | 0.937 | 0.930 |
|  | ±0.005 | ±0.005 | ±0.005 | ±0.005 |
| Causal Language | 0.863 | 0.827 | 0.863 | 0.827 |
|  | ±0.018 | ±0.020 | ±0.018 | ±0.020 |

Table A.3: Performance of our approach and baseline-TaskEmbed on the general and biomedical datasets. Both STL and MTL were fine-tuned using roberta-large.

| Task Group | Ours (Acc.) | Ours (F1) | TaskEmb (Acc.) | TaskEmb (F1) |
|---|---|---|---|---|
| THYMEMod | 0.958 | 0.743 | 0.958 | 0.740 |
|  | ±0.001 | ±0.007 | ±0.001 | ±0.008 |
| THYMENeg | 0.984 | 0.956 | 0.984 | 0.957 |
|  | ±0.001 | ±0.002 | ±0.001 | ±0.003 |
| SeedNeg | 0.986 | 0.955 | 0.982 | 0.943 |
|  | ±0.002 | ±0.005 | ±0.002 | ±0.006 |
| SeedUncert | 0.979 | 0.868 | 0.977 | 0.855 |
|  | ±0.005 | ±0.029 | ±0.003 | ±0.031 |
| StratNeg | 0.989 | 0.941 | 0.984 | 0.915 |
|  | ±0.001 | ±0.004 | ±0.002 | ±0.011 |
| StratUncert | 0.987 | 0.845 | 0.985 | 0.792 |
|  | ±0.002 | ±0.027 | ±0.002 | ±0.037 |

Table A.4: Performance of our approach and baseline-TaskEmbed on the clinical datasets. Both STL and MTL were fine-tuned using Bio+Clinical BERT.

| Task Group | Ours (Acc.) | Ours (F1) | Surrogate Models (Acc.) | Surrogate Models (F1) |
|---|---|---|---|---|
| BoolQ | 0.854 | 0.843 | 0.830 | 0.817 |
| | ±0.003 | ±0.004 | ±0.005 | ±0.005 |
| CB | 0.925 | 0.907 | 0.925 | 0.899 |
| | ±0.039 | ±0.056 | ±0.053 | ±0.090 |
| CB | 0.925 | 0.907 | 0.932 | 0.903 |
| | ±0.039 | ±0.056 | ±0.063 | ±0.100 |
| SST2 | 0.961 | 0.961 | 0.961 | 0.961 |
| | ±0.002 | ±0.002 | ±0.002 | ±0.002 |
| RTE | 0.842 | 0.841 | 0.657 | 0.590 |
| | ±0.015 | ±0.014 | ±0.180 | ±0.291 |
| COPA | 0.836 | 0.790 | 0.578 | 0.500 |
| | ±0.012 | ±0.018 | ±0.105 | ±0.197 |
| COPA | 0.836 | 0.790 | 0.611 | 0.544 |
| | ±0.012 | ±0.018 | ±0.126 | ±0.239 |
| BoolQ | 0.854 | 0.843 | 0.699 | 0.551 |
| | ±0.003 | ±0.004 | ±0.131 | ±0.285 |
| WiC | 0.686 | 0.679 | 0.540 | 0.433 |
| | ±0.017 | ±0.021 | ±0.076 | ±0.175 |
| COPA | 0.836 | 0.790 | 0.505 | 0.365 |
| | ±0.012 | ±0.018 | ±0.014 | ±0.090 |
| CoLA | 0.846 | 0.802 | 0.787 | 0.647 |
| | ±0.020 | ±0.027 | ±0.108 | ±0.270 |
| CB | 0.925 | 0.907 | 0.793 | 0.694 |
| | ±0.039 | ±0.056 | ±0.183 | ±0.290 |
| SST2 | 0.961 | 0.961 | 0.867 | 0.833 |
| | ±0.002 | ±0.002 | ±0.249 | ±0.344 |
| WiC | 0.686 | 0.679 | 0.625 | 0.587 |
| | ±0.017 | ±0.021 | ±0.087 | ±0.177 |
| Health Advice | 0.937 | 0.930 | 0.937 | 0.930 |
| | ±0.005 | ±0.005 | ±0.005 | ±0.005 |
| Causal Language | 0.863 | 0.827 | 0.863 | 0.827 |
| | ±0.018 | ±0.020 | ±0.018 | ±0.020 |

Table A.5: Performance of our approach and baseline-Surrogate Models on the general and biomedical datasets. Both STL and MTL were fine-tuned using roberta-large.

| Task Group | Ours (Acc.) | Ours (F1) | Surrogate Models (Acc.) | Surrogate Models (F1) |
| --- | --- | --- | --- | --- |
| THYMEMod | 0.958 | 0.743 | 0.958 | 0.740 |
| | ±0.001 | ±0.007 | ±0.001 | ±0.008 |
| THYMENeg | 0.984 | 0.956 | 0.984 | 0.957 |
| | ±0.001 | ±0.002 | ±0.001 | ±0.003 |
| SeedNeg | 0.986 | 0.955 | 0.983 | 0.944 |
| | ±0.002 | ±0.005 | ±0.004 | ±0.012 |
| THYMENeg | 0.984 | 0.956 | 0.984 | 0.955 |
| | ±0.001 | ±0.002 | ±0.001 | ±0.002 |
| SeedUncert | 0.979 | 0.868 | 0.977 | 0.859 |
| | ±0.005 | ±0.029 | ±0.006 | ±0.034 |
| StratUncert | 0.987 | 0.845 | 0.987 | 0.845 |
| | ±0.002 | ±0.027 | ±0.002 | ±0.027 |
| StratNeg | 0.989 | 0.941 | 0.984 | 0.915 |
| | ±0.001 | ±0.004 | ±0.002 | ±0.011 |
| StratUncert | 0.987 | 0.845 | 0.985 | 0.792 |
| | ±0.002 | ±0.027 | ±0.002 | ±0.037 |

Table A.6: Performance of our approach and baseline-Surrogate Models on the clinical datasets. Both STL and MTL were fine-tuned using Bio+Clinical BERT.

| Task Group | STL (Acc.) | STL (F1) | MTL (Acc.) | MTL (F1) |
| --- | --- | --- | --- | --- |
| BoolQ | 0.688 | 0.663 | 0.704 | 0.682 |
| | ±0.003 | ±0.005 | ±0.014 | ±0.015 |
| RTE | 0.612 | 0.606 | 0.632 | 0.626 |
| | ±0.026 | ±0.026 | ±0.040 | ±0.045 |
| CB | 0.682 | 0.477 | 0.861 | 0.836 |
| | ±0.009 | ±0.007 | ±0.019 | ±0.070 |
| WiC | 0.619 | 0.607 | 0.627 | 0.615 |
| | ±0.010 | ±0.012 | ±0.007 | ±0.009 |
| CoLA | 0.815 | 0.754 | 0.820 | 0.765 |
| | ±0.006 | ±0.008 | ±0.003 | ±0.007 |
| COPA | 0.526 | 0.524 | 0.539 | 0.534 |
| | ±0.047 | ±0.047 | ±0.023 | ±0.026 |
| SST2 | 0.922 | 0.922 | 0.921 | 0.921 |
| | ±0.005 | ±0.005 | ±0.008 | ±0.008 |
| CB | 0.682 | 0.477 | 0.839 | 0.760 |
| | ±0.009 | ±0.007 | ±0.057 | ±0.091 |
| Health Advice | 0.928 | 0.917 | 0.930 | 0.922 |
| | ±0.004 | ±0.005 | ±0.005 | ±0.006 |
| Causal Language | 0.823 | 0.750 | 0.860 | 0.829 |
| | ±0.016 | ±0.066 | ±0.008 | ±0.007 |

Table A.7: A comparison of performance between STL and MTL by task groupings with similar PVI estimates for the general and biomedical tasks on the test dataset with average result and a 95% confidence interval of 5 random seed values samples, where the t value for 95% confidence is 2.776. Both STL and MTL were fine-tuned using bert-base-uncased.

| Task Group | STL (Acc.) | STL (F1) | MTL (Acc.) | MTL (F1) |
|---|---|---|---|---|
| THYMENeg | 0.983 | 0.953 | 0.983 | 0.954 |
| | ±0.001 | ±0.002 | ±0.001 | ±0.002 |
| SeedNeg | 0.960 | 0.875 | 0.970 | 0.905 |
| | ±0.010 | ±0.024 | ±0.009 | ±0.027 |
| THYMEMod | 0.959 | 0.761 | 0.960 | 0.760 |
| | ±0.001 | ±0.009 | ±0.0001 | ±0.010 |
| SeedNeg | 0.960 | 0.875 | 0.961 | 0.881 |
| | ±0.010 | ±0.024 | ±0.009 | ±0.022 |
| SeedNeg | 0.960 | 0.875 | 0.962 | 0.881 |
| | ±0.010 | ±0.024 | ±0.008 | ±0.022 |
| StratNeg | 0.983 | 0.905 | 0.987 | 0.939 |
| | ±0.004 | ±0.020 | ±0.001 | ±0.009 |
| SeedUncert | 0.973 | 0.809 | 0.974 | 0.823 |
| | ±0.004 | ±0.025 | ±0.003 | ±0.014 |
| StratUncert | 0.980 | 0.589 | 0.985 | 0.775 |
| | ±0.002 | ±0.085 | ±0.003 | ±0.043 |

Table A.8: A comparison of performance between STL and MTL by task groupings with similar PVI estimates for the clinical tasks on the test dataset with average result and 95% confidence interval of 5 random seed values samples, where the t value for a 95% confidence is 2.776. Both STL and MTL were fine-tuned using BioBERT.

| Task Group | $T$-statistic | $P$-value |
|---|---|---|
| BoolQ-RTE | 1.372 | 0.186 |
| CB-CoLA | 1.841 | 0.082 |
| WiC-COPA | 0.921 | 0.362 |

Table A.9: Task groupings with PVI estimates that are not significantly different based on statistical measurements.

| Task Group | STL (Acc.) | STL (F1) | MTL (Acc.) | MTL (F1) |
|---|---|---|---|---|
| BoolQ | 0.731 | 0.707 | 0.744 | 0.723 |
| | ±0.028 | ±0.030 | ±0.020 | ±0.023 |
| RTE | 0.667 | 0.658 | 0.591 | 0.586 |
| | ±0.047 | ±0.056 | ±0.060 | ±0.064 |
| CB | 0.375 | 0.208 | 0.654 | 0.497 |
| | ±0.191 | ±0.101 | ±0.046 | ±0.094 |
| CoLA | 0.841 | 0.809 | 0.834 | 0.801 |
| | ±0.010 | ±0.015 | ±0.012 | ±0.012 |
| WiC | 0.670 | 0.668 | 0.675 | 0.674 |
| | ±0.021 | ±0.019 | ±0.025 | ±0.026 |
| COPA | 0.642 | 0.637 | 0.647 | 0.639 |
| | ±0.035 | ±0.043 | ±0.029 | ±0.038 |

Table A.10: A comparison of performance between STL and MTL by task groupings with similar PVI estimates for a sample of general domain tasks on the test dataset with average result and a 95% confidence interval of 5 random seed values samples, where the t value for 95% confidence is 2.776. Both STL and MTL were fine-tuned using meta-llama/Llama-3.2-1B model with LoRA.

| Task Group | Ours (Acc.) | Ours (F1) | MTL-all-task (Acc.) | MTL-all-task (F1) |
|---|---|---|---|---|
| BoolQ | 0.854 ±0.003 | 0.843 ±0.004 | 0.622 ±0.005 | 0.383 ±0.004 |
| CB | 0.925 ±0.039 | 0.907 ±0.056 | 0.482 ±0.005 | 0.216 ±0.002 |
| RTE | 0.842 ±0.015 | 0.841 ±0.014 | 0.527 ±0.000 | 0.345 ±0.000 |
| SST2 | 0.961 ±0.002 | 0.961 ±0.002 | 0.509 ±0.000 | 0.337 ±0.000 |
| COPA | 0.836 ±0.012 | 0.790 ±0.018 | 0.500 ±0.000 | 0.333 ±0.000 |
| WiC | 0.686 ±0.017 | 0.679 ±0.021 | 0.500 ±0.000 | 0.333 ±0.000 |
| CoLA | 0.846 ±0.020 | 0.802 ±0.027 | 0.691 ±0.000 | 0.409 ±0.000 |
| Health Advice | 0.937 ±0.005 | 0.930 ±0.005 | 0.589 ±0.000 | 0.247 ±0.000 |
| Causal Language | 0.863 ±0.018 | 0.827 ±0.020 | 0.458 ±0.000 | 0.157 ±0.000 |

Table A.11: Performance of our approach and baseline all tasks on the general and biomedical datasets. Both STL and MTL were fine-tuned using roberta-large.

| Task Group | Ours (Acc.) | Ours (F1) | MTL-all-task (Acc.) | MTL-all-task (F1) |
|---|---|---|---|---|
| THYMEMod | 0.958 ±0.001 | 0.743 ±0.007 | 0.958 ±0.001 | 0.740 ±0.009 |
| THYMENeg | 0.984 ±0.001 | 0.956 ±0.002 | 0.984 ±0.001 | 0.956 ±0.003 |
| SeedNeg | 0.986 ±0.002 | 0.955 ±0.005 | 0.983 ±0.002 | 0.944 ±0.001 |
| SeedUncert | 0.979 ±0.005 | 0.868 ±0.029 | 0.976 ±0.001 | 0.841 ±0.001 |
| StratNeg | 0.989 ±0.001 | 0.941 ±0.004 | 0.989 ±0.001 | 0.938 ±0.005 |
| StratUncert | 0.987 ±0.002 | 0.845 ±0.027 | 0.986 ±0.001 | 0.823 ±0.019 |

Table A.12: Performance of our approach and baseline all tasks on the clinical datasets. Both STL and MTL were fine-tuned using Bio+Clinical BERT.

| Task Group | Ours (Acc.) | Ours (F1) | MTL-random-task (Acc.) | MTL-random-task (F1) |
|---|---|---|---|---|
| BoolQ | 0.854 ±0.003 | 0.843 ±0.004 | 0.739 ±0.093 | 0.636 ±0.191 |
| CoLA | 0.846 ±0.020 | 0.802 ±0.027 | 0.771 ±0.066 | 0.622 ±0.171 |
| BoolQ | 0.854 ±0.003 | 0.843 ±0.004 | 0.728 ±0.091 | 0.623 ±0.180 |
| SST2 | 0.961 ±0.002 | 0.961 ±0.002 | 0.774 ±0.217 | 0.706 ±0.273 |
| COPA | 0.836 ±0.012 | 0.790 ±0.018 | 0.555 ±0.046 | 0.515 ±0.091 |
| Causal | 0.863 ±0.018 | 0.827 ±0.020 | 0.862 ±0.012 | 0.824 ±0.015 |
| CB | 0.925 ±0.039 | 0.907 ±0.056 | 0.821 ±0.035 | 0.733 ±0.100 |
| SST2 | 0.961 ±0.002 | 0.961 ±0.002 | 0.951 ±0.003 | 0.951 ±0.003 |
| Health Advice | 0.937 ±0.005 | 0.930 ±0.005 | 0.905 ±0.064 | 0.851 ±0.155 |
| RTE | 0.842 ±0.015 | 0.841 ±0.014 | 0.712 ±0.121 | 0.702 ±0.147 |
| CB | 0.925 ±0.039 | 0.907 ±0.056 | 0.840 ±0.084 | 0.770 ±0.153 |
| WiC | 0.686 ±0.017 | 0.679 ±0.021 | 0.665 ±0.018 | 0.661 ±0.018 |

Table A.13: Performance of our approach and baseline random tasks on the general and biomedical datasets. Both STL and MTL were fine-tuned using roberta-large.

| Task Group | Ours (Acc.) | Ours (F1) | MTL-random-task (Acc.) | MTL-random-task (F1) |
|---|---|---|---|---|
| THYMEMod | 0.958 ±0.001 | 0.743 ±0.007 | 0.958 ±0.002 | 0.739 ±0.011 |
| SeedUncert | 0.979 ±0.005 | 0.868 ±0.029 | 0.976 ±0.001 | 0.842 ±0.012 |
| StratNeg | 0.989 ±0.001 | 0.941 ±0.004 | 0.985 ±0.002 | 0.921 ±0.009 |
| THYMENeg | 0.984 ±0.001 | 0.956 ±0.002 | 0.983 ±0.001 | 0.953 ±0.001 |
| SeedNeg | 0.986 ±0.002 | 0.955 ±0.005 | 0.982 ±0.005 | 0.944 ±0.015 |
| StratUncert | 0.987 ±0.002 | 0.845 ±0.027 | 0.986 ±0.002 | 0.796 ±0.002 |

Table A.14: Performance of our approach and baseline random tasks on the clinical datasets. Both STL and MTL were fine-tuned using Bio+Clinical BERT.

| Setting | # Models | Trainable Parameters | Training Time | GPU Memory Consumption |
|---|---|---|---|---|
| Single Task Fine-tuning, all tasks | 15 | 15x355M= 5.325B | 15xT | ~20-24 GB x 15 (total) |
| Joint Learning, all tasks | 1 | 355M+15 task heads≈ 356M | ~1.5T- 2T | ~22-24 GB (total) |
| Joint Learning, pairwise | 7 | 7x(355M+2 task heads)≈ 2.665B | ~7x1.1T | ~22GB x 7 (total) |

Table A.15: A comparison of computational efficiency based on Roberta-large model on 15 tasks.
T = Baseline training time for one task (e.g., ~2 hrs on A40),
GPU Memory Consumption: Estimated peak memory during training using FP32 or mixed precision (AMP) on typical batch sizes

---

Sentence1: Then the silence in the Zoo became complete. Woil stared around him and then suddenly with a push of his wings raised himself into the air, turned, and landed ten feet away on the back of a green bench. Creggan could see that he was afraid and that his fear was making him terribly uncertain.
Sentence2: Woil was afraid
Question: Is this (0) entailment, (1) contradiction, or (2) neutral?
Answer: 0

Sentence1: He's weird enough to have undressed me without thinking, according to some mad notion of the "proper" thing to do. Perhaps he thought I couldn't lie in bed with my clothes on.
Sentence2: she couldn't lie in bed with her clothes on
Question: Is this (0) entailment, (1) contradiction, or (2) neutral?
Answer: 1

Sentence1: I hope you are settling down and the cat is well. This was a lie. She did not hope the cat was well.
Sentence2: the cat was well
Question: Is this (0) entailment, (1) contradiction, or (2) neutral?
Answer: 2

Table A.16: An example of a 3-shot prompt for the CB task. Our prompt design can be cross-validated with the approach in Chen et al. [62].

Context: Further research is needed to evaluate clinical applications of the PA, such as a more accurate identification of malnourished cardiac surgery patients.
Question: Is this (0) no relationship, (1) correlation, (2) conditional causation, or (3) direct causation?
Answer: 0

Context: The etiology of anemia appears to be iron-related and precipitated by the female sex.
Question: Is this (0) no relationship, (1) correlation, (2) conditional causation, or (3) direct causation?
Answer: 1

Context: Diet may influence the pharmacokinetics of ASA, but effects may be through modulation of glycine conjugation rather than glucuronidation.
Question: Is this (0) no relationship, (1) correlation, (2) conditional causation, or (3) direct causation?
Answer: 2

Context: Dietary interventions in older people were effective in maintaining fruit and fish intake, but this did not lead to a significant reduction in cognitive decline.
Question: Is this (0) no relationship, (1) correlation, (2) conditional causation, or (3) direct causation?
Answer: 3

Table A.17: An example of a 4-shot prompt for the Causal Language task.